\documentclass[discussion]{clv3}

\usepackage{url}            
\usepackage{booktabs}       
\usepackage{amsfonts}       
\usepackage{nicefrac}       
\usepackage{microtype}      
\usepackage{hyperref}
\usepackage{xcolor}
\usepackage{graphicx}
\usepackage{subfig}
\usepackage{cleveref}
\usepackage{wrapfig}
\usepackage{natbib}
\usepackage{arydshln}
\usepackage{multirow}
\usepackage{array}
\usepackage{makecell}
\usepackage[font=small,labelfont=bf]{caption}
\definecolor{darkblue}{rgb}{0, 0, 0.5}
\hypersetup{colorlinks=true,citecolor=darkblue, linkcolor=darkblue, urlcolor=darkblue}

\newcommand{\tabref}[1]{Table \ref{#1}}

\newcommand{\eqref}[1]{Eq.~\ref{#1}}

\newcommand{\secref}[1]{Section \ref{#1}}

\issue{xx}{x}{2020}

\runningtitle{Limitations of Stylometry for Neural Fake News Detection}

\runningauthor{Schuster et al.}

\begin{document}

\title{The Limitations of Stylometry for\\ Detecting Machine-Generated Fake News}

\author{Tal Schuster}
\affil{MIT, CSAIL \newline \texttt{tals@csail.mit.edu}}

\author{Roei Schuster}
\affil{Tel Aviv University, Cornell Tech \newline \texttt{rs864@cornell.edu}}

\author{Darsh J Shah}
\affil{MIT, CSAIL \newline \texttt{darsh@csail.mit.edu}}

\author{Regina Barzilay}
\affil{MIT, CSAIL \newline \texttt{regina@csail.mit.edu}}

\maketitle

\bibliographystyle{compling}

\begin{abstract}

Recent developments in neural language models (LMs) have raised concerns about their potential misuse for automatically spreading misinformation. In light of these concerns, several studies have proposed to detect machine-generated fake news by capturing their stylistic differences from human-written text. These approaches, broadly termed stylometry, have found success in source attribution and misinformation detection in human-written texts. However, in this work, we show that stylometry is limited against machine-generated misinformation. While humans speak differently when trying to deceive, LMs generate stylistically consistent text, regardless of underlying motive. Thus, though stylometry can successfully prevent impersonation by identifying text provenance, it fails to distinguish legitimate LM applications from those that introduce false information. We create two benchmarks demonstrating the stylistic similarity between malicious and legitimate uses of LMs, employed in auto-completion and editing-assistance settings.\footnote{Data: \url{https://people.csail.mit.edu/tals/publication/are_we_safe/}}
Our findings highlight the need for non-stylometry approaches in detecting machine-generated misinformation, and open up the discussion on the desired evaluation benchmarks.

\end{abstract}

\section{Introduction}
\label{sec:intro}

Many previous studies on stylometry---the extraction of stylistic features from written text---showed promising results on text classification. Two of stylometry's common applications are: (1) \textit{Detecting the provenance} of text (i.e.\ identifying the author) in order to prevent impersonations~\citep{Tweedie1996NeuralNA, brennan2012adversarial, Afroz2014DoppelgngerFT, anon15usenix, Neal2017SurveyingST, sari-etal-2017-continuous}; (2) \textit{Detecting misinformation} in text due to deception~\citep{Enos2007DetectingDU, Mihalcea2009TheLD, Ott2011FindingDO, Feng2012stylometry, Afroz2012DetectingHF}, fake news~\citep{rashkin2017truth, perez2017automatic}, or other false or illegal content~\citep{Choshen2019TheLO}. 
In the former, the classifier identifies language features that correlate with a specific person or group. The latter, misinformation detection, relies on idiosyncrasies of lies, i.e.\ style and language characteristics that are unique to text that is false or misleading.

Stylometry has recently gained attention as a potential answer to concerns that LMs could be used to mass-produce malicious text~\citep{Vosoughi1146, radford2019language}, that (1) impersonates a human author's text and/or (2) is fallacious and misleading.
Indeed, stylometry-based approaches have shown promising results for defending against human-impersonating LMs~\citep{bakhtin2019real, zellers2019defending}.
However, as applications of text generation such as text auto-completion~\citep{Wolf2019HuggingFacesTS} become widely used~\citep{I_lang_rob, Doc}, labeling text as generated by a LM might not indicate anything at all about its trustworthiness.
This motivates our core inquiry subject: \emph{can stylometry be used to distinguish malicious uses of LMs from legitimate ones?}

We build the first benchmark for detection of LM-produced fake news that labels text as ``real'' or ``fake'' according to its veracity.
Inspired by studies on deceitful behaviors in humans, showing that people try to diverge as little as possible from the truth when lying~\citep{mazar2008dishonesty}, we focus on automatic false modifications or additions to otherwise truthful news stories.

Our datasets contain articles produced by both malicious and responsible uses of language models, and the detector's task is to identify the malicious ones.
 In one dataset, we produce text by prompting a LM to extend news articles with relevant claims. We simulate malicious user, who only accepts the LM's suggestion if the claim is factually false, and a responsible user, who only accepts correct claims. The produced sentences are short and concise statements, similarly to fake news and false claims as represented in human-generated datasets~\citep{wang2017liar, Augenstein2019multifc}.
In another dataset, we modify existing news articles to include false information by inverting article statements. In this case, the LM is used to automatically identify the most plausible edit locations. This is similar to (mis-)using an autocorrect tool that suggests local modifications.

We find that with the state-of-the-art stylometry-based classifier, even a single auto-generated sentence within a wall of human-written text is detectable with high accuracy, yet the \emph{truthfulness} of a single sentence remains largely undecidable. Moreover, even a relatively weak LM can be used to produce statement inversions that the state-of-the-art stylometry-based model cannot detect. Thus, \emph{stylometry fails to distinguish malicious from responsible behaviors}. This indicates that, unlike humans who expose stylistic cues when writing false content~\citep{Ott2011FindingDO, frank2008human, matsumoto2011evaluating}, LMs maintain consistency for both true and false content. Worse yet, while a provenance classifier can effectively detect a potentially malicious author or publication venue (given enough examples), it might not distinguish malicious from legitimate authors if they are both using the same LM to generate their text. In this regard, \emph{malicious text generated by a LM might actually be harder to detect than hand-crafted malicious text}.

Our human evaluation tests show that humans are also fooled by machine-generated misinformation, but that access to external information sources can help. Therefore, we recommend future research on machine-generated misinformation to focus on non-stylometry strategies. Finally, we discuss what benchmarks are required for evaluating the performance of such detectors.
\section{Background and Related Work}


\paragraph{Stylometry for Human-written Text} 

The use of statistical methods for analyzing human-written documents has been studied extensively since the early days of the field. 
One common application is provenance detection. For example, \citet{authorship1963} used word counts to predict the authorship of historical documents. \citet{Tweedie1996NeuralNA} extracted other stylistic features and applied a neural network to the same task.
While these classifiers could be fooled by an aware writer that intentionally imitates other's style~\citep{brennan2012adversarial}, this approach was found useful for de-anonymizing cybercriminals in forums~\citep{Afroz2014DoppelgngerFT}, identifying programmers~\citep{anon15usenix} and more~\citep{Neal2017SurveyingST}.
In a related line of study, stylometry was applied to, rather than detecting the specific person, identifying characteristics of the author, such as age and gender~\citep{goswami2009stylometric}, political views~\citep{potthast-etal-2018-stylometric}, or IQ~\citep{abramov2019automatic}. Recently, as detailed later below, stylometry was used to distinguish machine- from human-writers.

Another common application of stylometry is detecting human-written misinformation. \citet{Mihalcea2009TheLD} found specific words that are highly correlated with true and false statements. \citet{Ott2011FindingDO, Feng2012stylometry} used a richer set of features such as POS tags frequencies and constituency structure to deceptive writing.
Following these observations and the increasing interest in fake news, recent studies applied stylometry on entire news articles~\citep{pisarevskaya-2017-deception}, short news reports~\citep{perez2017automatic}, fact and political statements~\citep{nakashole-mitchell-2014-language, rashkin2017truth} and posts in social media~\citep{volkova-etal-2017-separating}. The success of these studies is mostly attributed to stylistic changes in human language when lying or deceiving~\citep{bond2005prison,frank2008human}.
In this work, we evaluate the viability of this approach on machine-generated text, where stylistic differences between truth and lie might be more subtle.

\paragraph{Machine-generated Text Detection}
Detecting text's provenance is similar to authorship attribution and, therefore, stylometry can be effective. Indeed, \citet{gehrmann2019gltr} show the existence of distributional differences between human-written texts and machine-generated ones by visualizing the per-token probability according to a LM.
\citet{bakhtin2019real} learn a dedicated provenance neural classifier. While their classifier achieves high in-domain accuracy, they find that it overfits the generated text distribution rather than detecting outliers from human texts, resulting in increased ``human-ness'' scores for random perturbations.
Nevertheless, an advantage of such neural approaches over more traditional stylometry methods is that, given enough data, the model learns hidden stylistic representations without the need to manually define any features.

Building on the above observation, \citet{zellers2019defending} focus on fake news and create a Transformer-based LM~\citep{vaswani2017attention} dubbed Grover and train it on a large news corpus.
Grover also includes a ``neural fake news detector'', a linear classifier on top of the hidden state of the last token of the examined article, fine-tuned to classify if the news text was machine-generated or not. The experiments in this paper are based on the Grover-Mega classifier, fine-tuned for the target task (see section \ref{sec:adversarial_setting}).

\paragraph{Fake News Detection Approaches Beyond Stylometry} \label{back:alt}
The other most extensively studied NLP-based approach for fake news detection is based on fact-checking. This approach have recently gained increasing attention thanks to several synthetic~\citep{thorne2018fact} and real-world datasets~\citep{hanselowski2018retrospective, wang2017liar, popat-etal-2018-declare, Augenstein2019multifc}. The performance of current models is still far from humans~\citep{thorne-etal-2019-evaluating, schuster2019towards}, but with their advancements they can still play a positive role in detection.

Another line of work for fake news detection utilizes non-textual information such as how content is propagated, by which users, its originating URL, and other metadata~\citep{castillo2011information, gupta2014tweetcred, zhao2015enquiring, kochkina2018all, liu2018early}, as well as incorporating users' explicit feedback, such as abuse reporting~\cite{tschiatschek2018fake}. 
Social network platforms, ISPs, and even individual users can employ such methods to moderate content exposure.
These approaches are beset by the challenges~\cite{shu2017fake} of noisy and incomplete data, especially given the need to detect fake news early~\cite{liu2018early} (before propagation and user engagement patterns are fully formed).
\section{Adversarial Setting}
\label{sec:adversarial_setting}

\paragraph{``Fake News'': Our Working Definition} \label{par:truthfulness} 
Our attackers focus on automatically introducing false information into otherwise trustworthy content. We call the resultant manipulated articles ``fake news''. This definition matches the one of \citet{zhou2018fake} and is in line with the disinformation focused view of \citet{wardle2017information}.
Also, this is in line with how false claims are represented in many human-generated fake news datasets~\citep{wang2017liar, Augenstein2019multifc}.
Conversely, \citet{zellers2019defending} focus on entirely fabricated articles, a different type of fake news where the goal is mostly to create ``viral and persuasive'' content.

Our choice of creating articles with only a limited number of false statements is aligned with the way humans tend to deceive or lie. Psychological studies~\citep{mazar2008dishonesty} support the age-old notion that, when lying, 
\textit{``the best policy for the criminal is to tell the truth as nearly as possible''} [Raskolnikov, ``Crime and Punishment'' \cite{dostoyevsky2017crime}].
This helps preserve an honest self-image and, perhaps more importantly, reduces the chances of the lie being detected.
For example, a study on the longest-surviving known fake Wikipedia article~\citep{Wiki,Hox} revealed that many of the presented facts were only slightly altered from other, true facts.

\begin{figure}[t]
    \centering
    \small
     \includegraphics[width=1\textwidth]{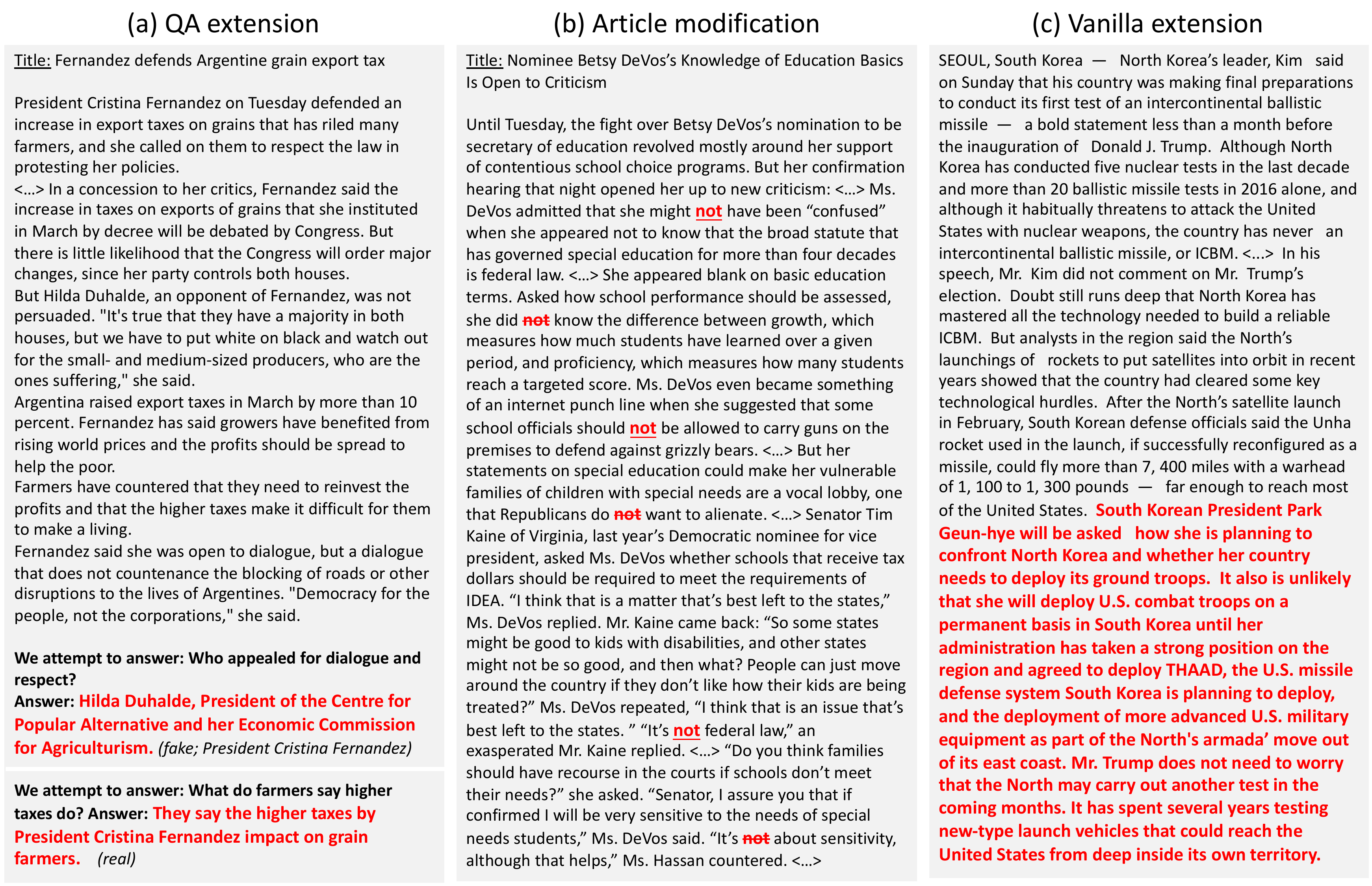}
      \vspace{-1\baselineskip}
     \caption{Examples of the fake class in our experiments. (a) In the news question answering (\secref{sec:close_exp}), a CNN article is presented with two examples of questions (bold) from newsQA~\citep{trischler-etal-2017-newsqa} and Grover's generated answer (red). The first answer is verified by a human annotator to be false and the second as true. (b) In article modification ($m=6$) (\secref{sec:close_exp}), the negations are marked with a cross-line for deletions and underline for addition. (c) In the article extension case (\secref{sec:hybridnews}), the bold red text is the generation of GPT-2 Medium to extend the prefix.}  \label{fig:examples}
    \vspace{-2\baselineskip}
\end{figure}

\paragraph{Attack and Defense Capabilities} We adopt an adversarial setting similar to that of \citet{zellers2019defending}. Our \textbf{attacker} wishes to generate \textit{fake} text, that contains unverified or false claims, en masse, using a language model to automate the process. The attacker's goal is to produce fake text that the verifier classifies as real.
Our \textbf{verifier} is \textbf{adaptive}: it receives a limited set of examples generated by the attacker, and trains a discriminator to detect the attacker's texts from legitimately-produced, \textit{real} text, containing exclusively human-verified claims (news articles from relatively reputable sources, like the The New York Times are assumed to be real).
We also experiment with a non-adaptive, \textbf{zero-shot} setting, where the verifier does not receive the attacker's examples.

\paragraph{Training and evaluating the detector}
\label{sec:esetup}
In each experiment, we collected a dataset with a ``real'' text class and a ``fake'' text class and used separate samples for testing and for fine-tuning. We used a Grover-Mega discriminator for all of the experiments. The model's weights were initialized from a checkpoint provided by \citet{zellers2019defending} and fine-tuned for 10 epochs with our training samples. 
For evaluating the zero-shot defense, we applied a pretrained Grover-Mega discriminator by querying its Web interface. 
We report human performance on some of the attacks.

\section{Stylometry Fails to Detect Machine-Generated Misinformation}
\label{sec:close_exp}


We create two datasets, simulating two different uses of LMs to automatically produce fake news. In the first, the \textit{extension scenario}, an auto-completion text generator extends a news article. A responsible user of this generator verifies the correctness of the output (producing real text), whereas an attacker verifies \emph{in}correctness (producing fake text).
In the second, the \textit{modification scenario}, the attacker uses a human-written news article and performs subtle modifications to semantically modify statements. Specifically, we add and remove negations. This follows the intuition that, if we take care to add negations in a syntactically-correct fashion, the new sentence is a negative inversion of the original~\citep{rudanko1982towards}. Yet, such changes are subtle enough to retain the original article's style and distributional features. See below for the full details on the creation of the datasets.


Additionally, we used about 100 examples from each dataset to test human performance in detecting this form of misinformation. For the extension scenario, we assigned two subject volunteers with two different tasks. The first had the same task as Grover's detector, and the second was allowed to use external sources to verify facts (which Grover cannot do). For the modification scenario, the participants had the same task as Grover but we highlighted negations to help them focus on the relevant parts.

\paragraph{(1) Creating the Extension Dataset}
\label{par:false_true_qa}
We use the newsQA dataset~\citep{trischler-etal-2017-newsqa} that contains Cable News Network (CNN) articles with corresponding questions and answers (segments copied verbatim from the text). Following \citet{radford2019language} and \citet{von2019training}, we embed the question in a template that is appended to the end of the input text:\\
\noindent\texttt{\small{} We attempt to answer: <question>\\
Answer:\\
}

\noindent Then, the first sentence produced by Grover's generator is used as a potential answer. We choose this template by examining common formats of questions in news articles and finding the one that empirically produces mostly reasonable answers by Grover's generator.
For example, given text about a suspect caught in Washington D.C., and the question \textit{``Where was the suspect stopped?''}, Grover answered \textit{``2 blocks from the U.S. Capitol.''}

We manually labeled the generated answers. We filtered out nonsensical ones (29\%) and labeled the rest as real or fake texts by correctness. Answers containing at least one false statement (judging by the article and/or external knowledge) were labeled fake.
We measured inter-annotator agreement on a subset of 100 examples labeled by two different annotators, and found it to be substantial (Cohen's kappa score of $k=0.78$).

Finally, we remove the sentence containing the answer from each article by removing the sentence with the highest TF-IDF-weighted word-count similarity with the question and answer (concatenated).

\paragraph{(2) Creating the Modifications Dataset}
We use articles from The New York Times~\citep{NYT}. We perform the modifications by removing and adding negations from statements: first, we randomly delete $m/2$ ``not'' or ``no'' occurrences from the text. Then, we automatically find statements to add $m/2$ negations to, using the probabilities of a GPT-2 Medium language model. We randomly sample 100 locations in the article and choose the ones with the maximal score, defined as the probability for either ``not'' or ``no'', multiplied by the probability for the word following the negation. By both adding and removing, we keep the total number of negations unchanged, avoiding anomalous negations counts. Original NYT articles are labeled as real and modified ones as fake. 

\subsection{Results}

\begin{table}[t]
\centering
\small
\begin{tabular}{clcccc}
\multicolumn{2}{l}{\textbf{Machine-generated misinformation}} & precision & recall & F1  & accuracy\\
\toprule
     
\multirow{4}{*}{\textbf{adaptive}}
 & QA extension (false vs.\ true) & 0.72  & 0.71 & 0.71 & 71\% \\
& modification ($m = 2$)&   0.53   & 0.52 & 0.53  & 53\% \\
& modification ($m = 6$)&  0.66    &  0.65    & 0.65  & 65\% \\
& modification ($m = 10$)&   0.73   &  0.47   & 0.63  & 65\% \\

\bottomrule\\
\end{tabular}
\vspace{-1\baselineskip}
	\caption{Results: \secref{sec:close_exp}. We report (macro) F1 score and overall accuracy, as well as precision and recall of the ``fake'' class. Zero-shot performance (not included) was very low in all cases.} \label{tab:close_exp}
	
\vspace{-2\baselineskip}
\end{table}

The detector's performance figures are given in \tabref{tab:close_exp}. Against QA extension, the detector reached 71\% accuracy in distinguishing true vs.\ false extensions generated by Grover. While this is above a trivial majority baseline (51\%), some of the success could be a side effect of the way that the dataset was constructed. Labeling instances with at least one false statement as fake makes these sentences generally longer, allowing a simple linear classifier that only uses the extension's length to reach 56\%.
When evaluated only on the short false answers (up to 10 words), the accuracy drops from 71\% to only 62\%.
Ultimately, however, despite performing better than simple baselines, the detection in this setting is much worse than in distinguishing human from machine text (\secref{sec:hybridnews}).

Comparing this with the results of the human evaluation is particularly instructive: the first subject, given the same task as the detector, attained an F1 score of 0.68, only marginally below the detector. The second participant, who was allowed to use external sources, attained an F1 of 0.84. 
We conclude that the stylometry-based classifier doesn't perform much better than humans in detecting potential misinformation, and that verifying against other resources can drastically improve results.

For the automatic article modification attack, the classifier fails completely to detect two modifications ($m=2$), and is not much better even when $m=10$ statements are inverted. 
Even though the change includes negations that might introduce surprising factual claims, the language style remains too similar to detect it. 
Humans, evaluated with $m=2$, had an F1 of 0.74, with 0.91 and 0.59 recall for real and modified articles, respectively. Thus, while humans perform better than the model, many of the modified statements were interpreted as real. Manually examining these, we found that the added negation actually changed the meaning in 60\% of the instances that were misclassified. 

Overall, the results show that both attacks can produce articles containing misinformation that evade the stylometry detector and mislead humans.
\section{Stylometry Detects Machine-Human Impersonations}
\label{sec:hybridnews}
\begin{table}[t]
\centering
\small
\begin{tabular}{clcccc}
\multicolumn{2}{l}{\textbf{Provenance detection}} & precision & recall & F1  & accuracy\\
\toprule
  \multirow{3}{*}{\textbf{zero-shot}}
&full article & 0.84      & 0.98   & 0.90 & 90\% \\
& vanilla extension ($g=20\%$) & 0.52   &   0.20 & 0.45 & 51\% \\
& vanilla extension ($g=1\%$) & 0.07    &  0.01 & 0.28 & 37\% \\
\midrule
\multirow{4}{*}{\textbf{\makecell{adaptive}}}
 & full article & 0.93 & 0.94 & 0.94 & 94\% \\
 & vanilla extension ($g=20\%$) & 0.90 & 0.97 & 0.95 & 95\% \\
 & vanilla extension ($g=1\%$) & 0.91 & 0.95 & 0.94 & 95\% \\
 & QA extension (machine vs.\ human) & 0.82 & 0.86 & 0.83 & 83\%  \\
\bottomrule\\
\end{tabular}
\vspace{-1\baselineskip}
	\caption{Detection results on the attacks of \secref{sec:hybridnews} in a zero-shot setting and the adaptive setting (where the discriminator is fine-tuned to a specific attacker). We report (macro) F1 score and overall accuracy. Precision, and recall of the ``fake'' class are reported as well.} \label{tab:farexp}

\vspace{-2\baselineskip}
\end{table}

One might suspect that the low performance in \secref{sec:close_exp} is due to limited capacity of our detector or the small fraction of generated text. We now show that the detector performs well on the provenance task and is sensitive to differences in small text portions.

\paragraph{(1) Fully-generated Articles}
We perform a first evaluation of Grover against texts generated by a different model of similar size, namely GPT-2 XL~\citep{radford2019language}. We include examples from the WebText test set~\citep{radford2019language}, labeled real text, and the released unconditioned generations of the GPT-2 XL model, labeled fake text.

\paragraph{(2) Automatic Article Extension (Vanilla)}
In order to experiment with mixed human-machine articles, we build an attacker that automatically extends unfinished human-written articles.
We simulate this attacker by querying a GPT-2 Medium model, conditioned on the the first 500 words from an article. We include New York Times articles, truncated to match the length of the fake class, labeled as real text, and articles that were automatically extended, labeled fake text. We use different percentages ($g$) of machine-generated text, by stopping the generation after an appropriate number of sentences.


\paragraph{(3) Automatic QA-based Extension}
In this experiment, we test whether the QA defense from \secref{par:false_true_qa} can perform better if we can assume that true texts are always human-written. To this end, we repeat the same setting but use the gold answers from the newsQA dataset for the ``real text'' class.



\subsection{Results}

\paragraph{Zero-shot Setting}
As \tabref{tab:farexp} shows, the zero-shot classifier is effective in detecting the fully generated articles of a different model, with a 0.9 F-1 score.
The extended articles, containing a substantial amount of human-written text, are mostly classified as human-written in this setting. This is unsurprising considering the dataset of full articles the detector was originally trained on.

\paragraph{Adaptive Setting}
After fine-tuning, Grover improves on the full articles and performs well against article extension generations, reaching 0.94 F1 score for articles with a single generated sentence ($g=1\%$). This stylometry detector is thus highly effective in distinguishing human from machine.


In the QA setting, also containing a single machine-generated sentence, the detector performs worse by 0.11 F1 points. This might be due to using the (stronger) Grover-Mega generator for the QA setting, and because the appended template for QA was selected to empirically maximize human ``reasonableness'' scores.
Still, this setting allows the stylometry detector to reach much higher performance than in the veracity-based setting (\secref{sec:close_exp}). This indicates that a restricted benchmark, that assumes no legitimate use of LMs, might not reflect the misinformation detection performance if the model is deployed in a world where LMs are used for both legitimate and malicious purposes.





\section{Discussion}
\label{sec:discussion}

Advancements in LM technology and their various applications have introduced a new challenge: distinguishing truthful text from misinformation, when the text is generated or edited by a LM. 
Our experiments indicate that LM-generated falsified texts are very similar in style to LM-generated texts containing true content. As a result, stylometry-based classifiers cannot identify auto-generated intentionally misleading content.

\noindent We conclude with the following recommendations:

\textit{(1) Extending Veracity-based Benchmarks.}
In order to better evaluate detectors against LM-generated misinformation, we recommend extending our benchmarks by creating other veracity-oriented datasets, 
that represent a wide range of LM applications, from whole-article generation to forms of hybrid writing and editing. 

\textit{(2) Improving Non-stylometry Methods.} Other detection approaches, as surveyed at the end of \secref{back:alt}, are less affected by the use of LMs. Therefore, advancements in such methods can improve the detection of both human- and machine-generated misinformation. Notably, the fact-checking setting makes fewer assumptions on the available auxiliary information and can be applied even if the text was sent to the verifier through a private channel such as E-mail. However, since fact-checking requires advanced inference capabilities, incorporating non-textual information, when available, can yield better results.

{}

\paragraph{Conclusion}
The potential use of LMs in creating fake news calls for a re-evaluation of current defense strategies. We examine the state-of-the-art stylometry model, and find it effective in preventing impersonation, but limited in detecting LM-generated misinformation.
This new kind of misinformation could be created by the same model that is used by legitimate writers as a writing-assistance tool, hiding stylistic differences between falsified and truthful content.
This motivates (1) constructing more instructive benchmarks for NLP-based approaches and improving non-stylistic methods, and (2) addressing a set of challenges that spans many disciplines beyond NLP, including social networks, information security, human-computer interaction, and others.

\paragraph{Acknowledgements}

We thank the anonymous reviewers and the members of the MIT NLP group for their helpful comments.
R.S is a member of the Check Point Institute of Information Technology.
This work is supported in part by Google's TensorFlow Research Cloud program;
DSO grant DSOCL18002; 
by the Blavatnik Interdisciplinary Cyber Research Center (ICRC); 
by the NSF award 1650589;
and by the generosity of Eric and Wendy Schmidt by recommendation of the Schmidt Futures program.



\starttwocolumn

\bibliography{references}

\end{document}